\documentclass[a4paper,conference]{IEEEtran}
\IEEEoverridecommandlockouts
\usepackage{cite}
\usepackage{amsmath,amssymb,amsfonts}
\usepackage{newtxtext,newtxmath}
\usepackage{graphicx}
\usepackage{textcomp}
\usepackage{xcolor}

\usepackage{booktabs}
\usepackage{array}
\usepackage{tabularx}
\usepackage{enumitem}
\usepackage{xurl}
\usepackage{tikz}
\usetikzlibrary{arrows.meta,positioning}
\usepackage{pgfplots}
\pgfplotsset{compat=1.16}
\usepackage[hidelinks]{hyperref}
\hypersetup{
  pdftitle={Semantic Signal-Assisted Inspection and Recovery Allocation in Reverse Logistics},
  pdfauthor={Jiani He, Dingyan Shang, Yihua Xu, Shiqi Huang, Yan Lyu, Jize Li, and Shangjing Tang},
  pdfkeywords={reverse logistics, text signals, inspection, recovery}
}

\newcolumntype{L}[1]{>{\raggedright\arraybackslash}p{#1}}
\newcolumntype{Y}{>{\raggedleft\arraybackslash}X}
\renewcommand\IEEEkeywordsname{Keywords}
\makeatletter
\newcommand{\authorrowbreak}{%
  \end{@IEEEauthorhalign}
  \hfill\mbox{}\par
  \mbox{}\hfill\begin{@IEEEauthorhalign}}
\makeatother

\begin{document}

\title{Semantic Signal-Assisted Inspection and Recovery Allocation in Reverse Logistics}

\author{
\IEEEauthorblockN{Jiani He\thanks{Corresponding author: Jiani He
(jianihe@alum.mit.edu). \textcopyright~2026 IEEE. Personal use of this material
is permitted. Permission from IEEE must be obtained for all other uses, in any
current or future media, including reprinting/republishing this material for
advertising or promotional purposes, creating new collective works, for resale
or redistribution to servers or lists, or reuse of any copyrighted component of
this work in other works.}}
\IEEEauthorblockA{\textit{Independent Researcher}\\
Seattle, USA\\jianihe@alum.mit.edu}
\and
\IEEEauthorblockN{Dingyan Shang}
\IEEEauthorblockA{\textit{Independent Researcher}\\
Frisco, USA\\dingyanshang@gmail.com}
\and
\IEEEauthorblockN{Yihua Xu}
\IEEEauthorblockA{\textit{Independent Researcher}\\
San Jose, USA\\yxu442@gmail.com}
\and
\IEEEauthorblockN{Shiqi Huang}
\IEEEauthorblockA{\textit{Independent Researcher}\\
Bellevue, USA\\juliahuangsq01@gmail.com}
\authorrowbreak
\IEEEauthorblockN{Yan Lyu}
\IEEEauthorblockA{\textit{Independent Researcher}\\
Boston, USA\\lyu.yan@northeastern.edu}
\and
\IEEEauthorblockN{Jize Li}
\IEEEauthorblockA{\textit{Independent Researcher}\\
Boston, USA\\jizel@bu.edu}
\and
\IEEEauthorblockN{Shangjing Tang}
\IEEEauthorblockA{\textit{Independent Researcher}\\
Indianapolis, USA\\tangshangjing@gmail.com}}

\maketitle

\begin{abstract}
Reverse-logistics operators often decide how to inspect and route returned assets
before their condition is fully observed, while full inspection consumes scarce
labor. Semantic Signal-Assisted Decision Support converts return notes into a
condition factor and a signal-quality score that guide inspection depth and
recovery allocation under shared labor capacity. We evaluate the framework in
three synthetic benchmark scenarios spanning information technology decommissioning,
aircraft maintenance, and consumer-electronics returns. Across 30 paired simulation
seeds, the keyword implementation improves net recovery value relative to a
structured-feature comparator with noisy full inspection while reducing inspection
cost in all three scenarios. A risk-blind comparator that skips inspection
altogether still records higher value under the benchmark's purely economic objective. At
matched inspection cost, score-guided targeting
adds 53.9 thousand United States dollars per batch in the aircraft scenario but has
little economic effect in the other two configurations; phrase and large language
model extractors provide further gains in the aircraft scenario. These results show
how narrative evidence can support inspection allocation before recovery decisions
are made.
\end{abstract}

\begin{IEEEkeywords}
reverse logistics, text signals, inspection, recovery
\end{IEEEkeywords}

\section{Introduction}\label{sec:intro}

Resource recovery through reverse logistics is a large and growing operational
challenge. The global data-center information technology (IT) asset disposition market was estimated at \$13.1 billion in
2025~\cite{GlobalMktInsights}; aircraft maintenance, repair, and overhaul (MRO)
demand was forecast at \$104 billion for 2024~\cite{OliverWyman}, and United States
retail returns were projected to reach \$890 billion, or 16.9\% of annual sales, in 2024~\cite{NRF}.

The practical bottleneck is condition uncertainty. Full inspection reduces this
uncertainty, but it requires time, labor, and sometimes specialized equipment. Many
assets already arrive with narrative evidence, such as technician notes, maintenance
records, and customer descriptions, but standard reverse-logistics optimization
models expect structured inputs such as prices, disassembly costs, yield priors,
and capacity, not free-text annotations~\cite{Slama2022,Govindan2015,Sun2024}.

To use this information in a decision model, Semantic Signal-Assisted Decision
Support (SSADS) maps each return note to a condition factor and a signal-quality score.
The note is treated as noisy evidence
available before inspection. The condition factor shifts expected recovery yield,
and the signal-quality score sets inspection depth:
an asset may skip discretionary inspection, receive a quick functional test, or undergo full
component-level inspection. The recovery optimizer then ranks assets by expected net
value under capacity constraints and assigns each asset to a scenario-feasible
recovery disposition or scrap. In this formulation, inspection becomes an asset-level allocation
decision rather than a uniform preprocessing step. We evaluate SSADS using the
Reverse Logistics Decision Benchmark (RLDB) across
three structurally distinct scenarios to test how return notes can affect recovery
decisions. The scenarios use descriptive technician notes to target inspection,
maintenance records to support inspection prioritization in a regulated workflow,
and short customer descriptions to triage high-volume returns under limited
recovery capacity.

\begin{itemize}[leftmargin=*,noitemsep]
\item \textbf{S1 (IT Infrastructure):} Planned IT decommissioning with descriptive
  technician notes and minimal regulatory constraints.

\item \textbf{S2 (Aircraft MRO):} The repair-station setting under Part~145 of
  Title~14 of the Code of Federal Regulations (14~CFR Part~145), where
  note-derived scores prioritize work within the inspection and return-to-service
  controls required by the station's quality
  system~\cite{eCFRPart145}.

\item \textbf{S3 (Consumer Electronics):} Consumer returns emphasize scale.
  Customer text is noisier and unit value is lower, making triage the central task.
\end{itemize}

This paper makes three contributions:
\begin{enumerate}[noitemsep]
\item \textbf{Inspection targeting from return notes:} We formulate return-note use
  as an inspection-allocation problem. SSADS maps each note to a condition factor
  $\phi$ and a signal-quality score
  $\sigma$, so expected yield and inspection depth can be adjusted at the asset
  level.

\item \textbf{Modular recovery pipeline:} SSADS keeps the text reader separate
  from the recovery optimizer. As long as a reader returns $\phi$ and $\sigma$,
  the same inspection policy and capacity-constrained allocator can be used with
  keyword rules, phrase matching, or large language model extraction.

\item \textbf{Reproducible benchmark evaluation:} We use RLDB to test the same mechanism across IT
  decommissioning, aircraft MRO, and consumer-electronics
  returns. The benchmark uses shared baselines, paired seeds, and a three-level
  extractor ladder to measure how extractor accuracy translates into inspection
  targeting and recovery value. We additionally validate the allocator against the
  exact optimum of the corresponding fixed-action 0--1 admission problem and
  compare equal-cost inspection policies.
\end{enumerate}

Section~II positions SSADS in prior work, Section~III defines the framework,
Section~IV reports the benchmark, and Sections~V--VI discuss implications and
conclude.

\section{Background and Related Work}

Reverse logistics decisions are sequential and partially irreversible.
Optimization work spans stochastic disassembly lot-sizing~\cite{Slama2022},
broader reverse-logistics and closed-loop supply-chain design~\cite{Govindan2015},
and recent simulation--optimization frameworks for dynamic reverse-logistics
network design~\cite{Sun2024},
but it generally requires structured numerical inputs: disassembly bills of materials
(BOM)~\cite{Babbitt2020}, cost tables, and yield rates specified before optimization.
A parallel line predicts condition or remaining useful life (RUL) from sensor and
maintenance data, from run-to-failure prognostics~\cite{Saxena2008} to deep RUL
models~\cite{Zhang2017}; disassembly research also optimizes processing
sequences~\cite{Fan2023}. These methods address prognosis or sequencing rather than
jointly choosing text-guided inspection depth and recovery allocation. In practice,
much decision-relevant information lives in unstructured narratives
that existing optimization models in this setting generally do not ingest without
manual translation; information-extraction schemas for maintenance text~\cite{Bikaun2024}
structure the narrative but stop short of the disposition decision.

Recent text-based methods have advanced in forward supply chains: OptiGuide~\cite{Li2023}
translates language into optimization code, InvAgent~\cite{Quan2024} coordinates
inventory agents, and others extract supply-chain structure via zero-shot learning or
large language models (LLMs)~\cite{AlMahri2026,Liu2024}. These cited systems target forward-chain
optimization or structure extraction rather than reverse-logistics disposition under
yield uncertainty.

Maintenance research has also begun to connect narrative records with operational
decisions. Recent work automates the analysis and assignment of maintenance work
orders~\cite{LiMWO2024} and extracts causal relations from long maintenance
documents~\cite{Hershowitz2024}. Deng \emph{et al.} use an LLM agent for context-aware
maintenance decision support~\cite{Deng2024}, while Getz and Tong use LLMs to
accelerate maintenance insight generation~\cite{Getz2025}. SSADS instead maps narrative
evidence to inspection depth and recovery allocation under shared labor capacity.

Using an extractor's signal-quality score to set inspection depth is motivated by
value-of-information reasoning~\cite{Howard1966}: reserve costly measurements for
less informative records. Similar logic appears in active learning~\cite{Settles2009}
and optimal-inspection work~\cite{Wang2002}. SSADS uses fixed score thresholds rather
than explicitly estimating the value of information, and its interface is independent
of the text extractor.

\section{Framework Design}

\subsection{Problem Setting}

Assets arrive in periodic batches. Each asset $m$ has structured attributes
(type, age, and bill of materials) and a return note $\mathcal{T}(m)$. SSADS
selects an inspection level $q_m\in\{0,1,2\}$ (skip, quick, or full) before
selecting a scenario-feasible recovery action or scrap. For a component-recovery action $a$,
the expected gross value is
\begin{equation}
  G(m,a) = \sum_{c\in\mathrm{BOM}(m,a)} n_{mc}p_c\,\tilde y_c(m),
  \label{eq:value}
\end{equation}
where $n_{mc}$ is component count, $p_c$ is recovery price, and
$\tilde y_c(m)$ is the current expected yield. Whole-unit refurbishment instead
uses its configured asset-level value times $\phi(m)$. A partial-recovery teardown
recovers a configured fraction (0.60 in all three scenarios) of the component
value at lower cost and time; it is reachable by the routing comparators, whereas
the margin-ranked allocator chooses between component recovery and whole-unit
refurbishment. The processing margin
subtracts action cost and compares each recovery action with the default scrap disposition.
Inspection cost and time are always charged, including for assets later scrapped
because capacity is exhausted; scrap handling also consumes cost and time.

\subsection{System Architecture}

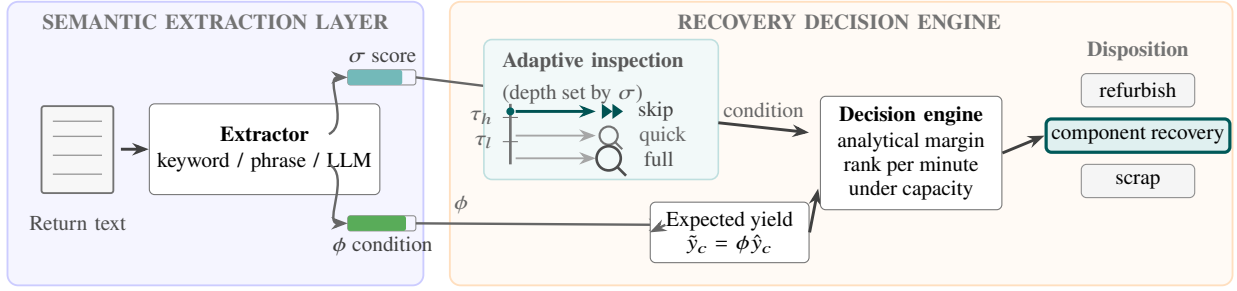
\begin{figure*}[t]
  \centering
  \begin{tikzpicture}[
      x=1cm, y=1cm, font=\footnotesize,
      >={Stealth[length=2.2mm]},
      proc/.style={rounded corners=2pt, draw=black!55, fill=white,
      align=center, inner sep=3pt},
      chip/.style={rounded corners=2pt, draw=black!45, fill=black!4,
        align=center, font=\footnotesize, minimum width=15mm,
      minimum height=4.4mm, inner sep=2pt},
      ptitle/.style={font=\footnotesize\bfseries, text=black!55},
      flow/.style={->, line width=1.1pt, draw=black!75},
      sig/.style={->, line width=0.9pt, draw=black!60},
    faint/.style={->, line width=0.7pt, draw=black!35}]

    \fill[blue!4, rounded corners=4pt]   (0,0.2)    rectangle (5.5,3.95);
    \draw[blue!20, rounded corners=4pt, line width=0.7pt] (0,0.2) rectangle (5.5,3.95);
    \node[ptitle] at (2.75,3.68) {SEMANTIC EXTRACTION LAYER};
    \fill[orange!4, rounded corners=4pt] (5.85,0.2) rectangle (16.2,3.95);
    \draw[orange!25, rounded corners=4pt, line width=0.7pt] (5.85,0.2) rectangle (16.2,3.95);
    \node[ptitle] at (11.0,3.68) {RECOVERY DECISION ENGINE};

    \begin{scope}[shift={(0.45,1.42)}]
      \draw[rounded corners=1pt, fill=black!3, draw=black!55, line width=0.7pt]
      (0,0) rectangle (0.95,1.15);
      \foreach \yy in {0.25,0.5,0.75,0.95}
      \draw[black!40, line width=0.5pt] (0.15,\yy) -- (0.8,\yy);
    \end{scope}
    \node[font=\footnotesize, align=center, text=black!70] at (0.93,1.02)
    {Return text};

    \node[proc, minimum width=28mm, minimum height=12mm] (ext) at (3.4,2.0)
    {\textbf{Extractor}\\[1pt]{\footnotesize keyword / phrase / LLM}};
    \draw[flow] (1.5,2.0) -- (ext.west);

    \node[font=\footnotesize, text=black!75] at (4.95,3.22) {$\sigma$ score};
    \draw[draw=black!45, rounded corners=1pt] (4.5,2.85) rectangle (5.4,3.05);
    \fill[teal!55, rounded corners=1pt]       (4.5,2.85) rectangle (5.22,3.05);
    \node[font=\footnotesize, text=black!75] at (4.95,0.72) {$\phi$ condition};
    \draw[draw=black!45, rounded corners=1pt] (4.5,0.92) rectangle (5.4,1.12);
    \fill[green!50!black!65, rounded corners=1pt] (4.5,0.92) rectangle (5.27,1.12);
    \draw[sig] (4.25,2.2) to[out=25,in=180]  (4.5,2.95);
    \draw[sig] (4.25,1.8) to[out=-25,in=180] (4.5,1.02);

    \draw[sig] (5.4,2.95) -- (6.6,2.8);                       
    \draw[sig] (5.4,1.02) -- (8.7,1.0)
    node[pos=0.18, above, font=\footnotesize, text=black!60] {$\phi$}; 

    \fill[teal!5, rounded corners=3pt] (6.3,1.6) rectangle (9.4,3.45);
    \draw[teal!25, rounded corners=3pt, line width=0.6pt] (6.3,1.6) rectangle (9.4,3.45);
    \node[anchor=north west, align=left, font=\footnotesize, text=black!65] at (6.42,3.4)
    {\textbf{Adaptive inspection}\\[3pt](depth set by $\sigma$)};
    \draw[black!55, line width=1pt] (6.65,1.8) -- (6.65,2.6);    
    \draw[black!55] (6.55,2.42) -- (6.75,2.42);
    \node[font=\footnotesize, text=black!60, left=0pt] at (6.55,2.42) {$\tau_h$};
    \draw[black!55] (6.55,2.1) -- (6.75,2.1);
    \node[font=\footnotesize, text=black!60, left=0pt] at (6.55,2.1) {$\tau_l$};
    \fill[teal!75!black] (6.65,2.5) circle (1.5pt);               
    \draw[flow, draw=teal!65!black] (6.72,2.5) -- (7.78,2.5);
    \draw[faint] (6.72,2.18) -- (7.78,2.18);
    \draw[faint] (6.72,1.88) -- (7.78,1.88);
    \fill[teal!65!black] (7.86,2.4)--(7.86,2.6)--(8.0,2.5)--cycle;  
    \fill[teal!65!black] (8.0,2.4)--(8.0,2.6)--(8.14,2.5)--cycle;
    \node[font=\footnotesize, text=black!75, right=1pt] at (8.18,2.5) {skip};
    \draw[black!55, line width=0.8pt] (7.95,2.18) circle (0.12);    
    \draw[black!55, line width=0.8pt] (8.03,2.1) -- (8.13,2.0);
    \node[font=\footnotesize, text=black!60, right=1pt] at (8.2,2.18) {quick};
    \draw[black!70, line width=1pt] (7.95,1.88) circle (0.16);     
    \draw[black!70, line width=1pt] (8.06,1.77) -- (8.18,1.65);
    \node[font=\footnotesize, text=black!70, right=1pt] at (8.26,1.87) {full};
    \draw[flow] (9.4,2.35) -- (10.6,2.22)
    node[pos=0.5, above, font=\footnotesize, text=black!60] {condition};

    \node[proc, minimum width=21mm, minimum height=8mm] (yield) at (9.55,0.9)
    {Expected yield\\$\tilde y_c=\phi\hat y_c$};
    \draw[sig] (8.7,1.0) -- (yield.west);
    \draw[flow] (yield.east) -- (10.72,1.48);

    \node[proc, minimum width=24mm, minimum height=15mm] (eng) at (11.95,1.95)
    {\textbf{Decision engine}\\[1pt]{\footnotesize analytical margin}\\
    {\footnotesize rank per minute}\\{\footnotesize under capacity}};

    \node[ptitle] at (14.95,3.3) {Disposition};
    \node[chip] at (14.95,2.78) {refurbish};
    \node[chip, draw=teal!70!black, fill=teal!12, very thick] (sel) at (14.95,2.18) {component recovery};
    \node[chip] at (14.95,1.58) {scrap};
    \draw[flow] (eng.east) -- (sel.west);
  \end{tikzpicture}
  \caption{The SSADS pipeline. The \emph{semantic extraction layer} reads return
    text into two scalars: a condition factor $\phi$ (the yield-prior shift) and a
    signal-quality score $\sigma$. In the \emph{recovery decision engine}, $\sigma$ gates
    inspection depth (skip when $\sigma{\ge}\tau_h$, quick functional test, or full
    component inspection when $\sigma{<}\tau_l$), while $\phi$ shifts expected yield.
    After inspection, an analytical expected-margin step ranks assets under the
    remaining shared labor capacity and selects a disposition (highlighted:
    component recovery).}
  \label{fig:pipeline}
\end{figure*}

SSADS has two layers (Fig.~\ref{fig:pipeline}). The \emph{Semantic Extraction
Layer} reads each note and returns $\phi\in(0,1]$ and $\sigma\in[0,1]$. The
\emph{Recovery Decision Engine} uses $\phi$ to set the yield prior and $\sigma$ to
gate inspection depth. After every selected inspection has been performed and
charged, the engine ranks positive expected margins by value per processing minute
and allocates the remaining shared labor capacity.

\subsection{Semantic Extraction Layer}
Any extractor that provides a condition factor $\phi\in(0,1]$ and a signal-quality score
$\sigma\in[0,1]$ can be used without changing the inspection policy or decision
engine. This lets the keyword, phrase matcher, and LLM readers be compared under
the same downstream decision logic. The main benchmark uses a restricted-vocabulary
keyword-and-pattern classifier, which falls back to $\phi{=}1.0$ when no signal
triggers. The phrase matcher and cached LLM provide stronger comparison points in
the end-to-end recovery evaluation (Section~\ref{sec:cal2val}). All extractors
receive text alone. Under the configured note-generation noise
($p_\text{omit}{=}0.15$, $p_\text{mislabel}{=}0.25$), their reported correlations
remain below $r{=}1$. Negative signals lower $\phi$ multiplicatively, while positive
signals cannot increase it beyond $1.0$.

The phrase matcher is deterministic and scenario-specific. It counts declared
phrases for each condition, chooses the condition with the largest count, and maps
that condition to a fixed $\phi$ and $\sigma{=}0.90$; declared condition order breaks
ties. If no phrase matches, it returns $\phi{=}1.0$, $\sigma{=}0.20$, and a fallback
flag. The complete phrase lists and maps are in
\path{experiments/src/s2s/extractors/strong.py}; no generated template produces a
cross-condition top-score tie.

\smallskip\noindent\textbf{Prompted large language model extractor.}
The LLM rung uses the DeepSeek chat-completions application programming interface
(API)~\cite{DeepSeekAPI} with the
model identifier \texttt{deepseek-chat} and temperature $0$. A scenario-specific system
prompt, set before any note is read, specifies the $\phi$ rubric (e.g.\ S1:
${\sim}0.9$ clean, ${\sim}0.6$ mixed, ${\sim}0.2$ damaged, $0.5$ uninformative) and
asks for a separate score $\sigma$ for how much condition information the note
contains, independently of $\phi$. The prompt requests a JavaScript Object Notation (JSON)
object with fields \texttt{\{phi,sigma,condition\}}, where \texttt{phi} and
\texttt{sigma} encode $\phi$ and $\sigma$. It uses the model's API JSON mode; we
parse and validate the scores against $(0,1]\times[0,1]$
(\path{experiments/src/s2s/extractors/deepseek.py}).
The author-defined rubric maps free text onto a specified condition scale; recovery
value is computed downstream by the decision engine.

\subsection{Yield Adjustment and Inspection}\label{sec:prior}

Given $\phi(m)$ and component baseline yield $\hat{y}_c$, the pre-inspection
expected yield is
\begin{equation}
  \tilde y_c(m)=\phi(m)\hat y_c.
  \label{eq:prior}
\end{equation}
This analytical mean is sufficient because the allocation objective is linear and
has no recourse. The simulator still draws realized yields from a Beta distribution
with fixed concentration 20, centered on the component baseline yield multiplied by
the latent condition factor, when computing realized net value. A score-dependent
concentration parameter is omitted because it has no decision effect under the
present mean-value objective.

\subsection{Adaptive Inspection Policy}

The system uses signal-quality score $\sigma(m)$ to determine inspection depth for each asset:
\begin{itemize}[noitemsep]
  \item $\sigma\geq\tau_h$: rely on $\phi$ without additional inspection.
  \item $\tau_l\leq\sigma<\tau_h$: run a quick functional test.
  \item $\sigma<\tau_l$: use full component-level inspection.
\end{itemize}
For S1 and S2, the prespecified thresholds are $\tau_h{=}0.5$ and
$\tau_l{=}0.25$; S3 uses
$\tau_l{=}0.45$ because colloquial text rarely gets high scores.
A quick or full inspection gives a noisy condition observation with standard
deviation 0.15 or 0.05, respectively. The observation receives 50\% or 90\%
weight in the updated condition factor. Inspection happens before allocation, so
its cost and time remain charged even if an asset is not processed. A skipped
inspection saves \$20--\$100 and 5--60 minutes per asset, but can leave an
over-optimistic recovery estimate. These thresholds are prespecified reference
settings rather than claimed recovery-value optima;
Section~\ref{sec:sensitivity} reports results for nearby settings.

\subsection{Decision Engine}

The engine processes weekly batches, except in S3 where batches are daily. Every
asset has scrap cost $C_s$ and time $t_s$. For asset $m$, let $\Delta M_m$ be the
best positive expected margin of a recovery action relative to scrap, and let
$\Delta t_m$ be its incremental time. Let $x_m\in\{0,1\}$ indicate whether that
action replaces scrap. With $H$ total labor minutes and inspection time $h_m(q_m)$,
the allocator approximately solves
\begin{equation}
 \begin{aligned}
 \max_{x_m\in\{0,1\}}\;&\sum_m \Delta M_mx_m \\
 \mathrm{s.t.}\;&\sum_m \Delta t_mx_m\le H-\sum_m[h_m(q_m)+t_s].
 \end{aligned}
 \label{eq:allocation}
\end{equation}
It sorts candidates by $\Delta M_m/\Delta t_m$ and admits them while capacity remains. We compare
this heuristic with the exact optimum of the fixed-action 0--1 admission problem on
every full 500/1,000-asset benchmark instance, solved with SciPy's mixed-integer solver~\cite{Virtanen2020} whenever candidate
action times differ. In RLDB every candidate shares a single processing time once
its best action is fixed, so the exact optimum reduces to admitting the
highest-margin feasible assets; the mean and maximum expected-objective gaps are
0.0\% in all three scenarios. The scenario parameters used by this allocation
are summarized in Table~\ref{tab:scenarios}.

\begin{samepage}
\smallskip\noindent\textbf{Batch allocation summary.}
For each asset, (1)~extract $(\phi,\sigma)$ from its return note;
(2)~select skip, quick, or full inspection from the score thresholds and update
the condition estimate when inspection occurs; (3)~charge inspection and default
scrap handling; (4)~compute each feasible action's expected margin and processing
time; and (5)~rank positive-margin candidates by margin per minute and admit them
while the remaining labor capacity permits. The output is one disposition per
asset together with inspection, cost, time, and realized-value records.
\end{samepage}

\subsection{Illustrative Asset Walkthrough}
The following asset illustrates the decision path. A decommissioned server with note ``Routine decommission.
All components seated properly. No corrosion. 4yr service.'' is read as a healthy unit
($\phi{=}0.925$, $\sigma{=}0.925$; \texttt{run\_diagnostics.py}). Since $\sigma$
is above $\tau_h{=}0.5$, the policy skips inspection and, using \eqref{eq:prior}, ranks
component recovery as the highest priority. This saves the \$75 and 60-minute
inspection required by a full-inspection policy. A second note refers to the power
supply unit (PSU) and central processing unit (CPU): ``PSU failure. Visible burn
marks on mainboard near power connector J12. CPU smells burnt.'' It returns
$\phi{=}0.400$ and $\sigma{=}0.875$.
Its high $\sigma$ still skips inspection, while its lower $\phi$ reduces expected value
and allocation priority without necessarily implying scrap. Section~\ref{sec:error} examines
cases in which high-score text overestimates condition.

\section{Cross-Industry Evaluation}

\subsection{Scenarios, Data, and Baselines}

\begin{table}[!t]
\caption{Evaluation Scenarios}
\label{tab:scenarios}
\centering\footnotesize
\setlength{\tabcolsep}{1pt}
\begin{tabular}{@{}lccc@{}}
\toprule
 & S1: IT & S2: Aircraft & S3: Consumer \\
\midrule
Scenario type  & Product       & Service        & Product     \\
Return driver  & Decommission  & Degradation    & Returns     \\
Core uncertainty & Component yield & Condition & Condition \\
Text source    & Technician notes & Maint. records & Customer desc. \\
Full inspection & \$75, 60 min  & \$100, 35 min  & \$20, 5 min \\
Quick inspection & \$19, 15 min & \$50, 20 min & \$10, 3 min \\
Labor capacity & 600 h/week & 800 h/week & 160 h/day \\
Assets/batch   & 500           & 500            & 1,000       \\
Regulation     & Minimal       & \begin{tabular}[t]{@{}c@{}}14 CFR\\Part 145\end{tabular}
               & \begin{tabular}[t]{@{}c@{}}Consumer\\safety\end{tabular} \\[2pt]
\bottomrule
\end{tabular}
\end{table}

\smallskip\noindent\textbf{Noisy note generation.}
Latent condition sets true yield and anchors a separately corrupted textual
observation.
We corrupt notes two ways: omission ($p{=}0.15$, the note becomes uninformative) and
a one-class severity-perturbation attempt ($p{=}0.25$). At an endpoint, outward
perturbations are clipped and can leave the observed class unchanged. Thus, text has
an intentional but imperfect link to yield
(\path{experiments/src/data_generators/noise.py}). The corpus uses 17/20/21
condition-class base templates for S1/S2/S3, with S2 written in Service Difficulty
Report style. Numeric age,
station, and hour fields yield 27/4{,}840/43 unique rendered note strings over seeds 0--29.
Mean note lengths are 9.8/9.6/8.0 words, with vocabularies of 133/3,690/152
tokens and 4/6/5 true-condition classes for S1/S2/S3.
Additional note profiles, scenario parameters, and LLM cache coverage are available
in the companion repository~\cite{RLDBRepo}.

\smallskip\noindent\textbf{Baselines.}
We compare seven alternatives with the same generated assets and seed.
(1)~\emph{Random routing}: no inspection and a random choice among partial
recovery, component recovery, and scrap, subject to the same capacity accounting.
(2)~\emph{First-in, first-out (FIFO) routing}: no inspection and processing in
arrival order.
(3)~\emph{Structured + noisy full inspection}: a gradient-boosting regressor trained on an independent
4,000-asset population (seed 99999) using age, asset type, and BOM component counts,
followed by noisy full inspection and the common allocator.
(4)~\emph{Structured + semantic adaptive inspection}: the same structured features plus keyword $\phi$, $\sigma$, and
fallback status, followed by adaptive inspection and the common allocator.
(5)~\emph{Oracle full inspection}: exact condition revelation through full inspection, followed by
the common allocator.
(6)~\emph{No-signal / no-inspection}: no condition signal, no inspection, and the common allocator.
This risk-blind comparator pursues the benchmark's economic objective without
paying for information.
(7)~\emph{Semantic-only / threshold routing}: text-guided adaptive inspection with
fixed threshold routing in place of the margin-ranked allocator. SSADS instead pairs
the same semantic inspection policy with the capacity-constrained value allocator.
After noisy full inspection, the structured
baseline updates $\phi$ with 90\% weight on a clipped Gaussian observation and 10\%
on its structured pre-inspection estimate. The oracle instead replaces $\phi$ with
the exact latent yield factor; both then invoke the same allocator.

\smallskip\noindent\textbf{Metrics.}
\emph{Total recovery value} (TRV) is realized gross recovery less processing,
inspection, post-allocation rework, and disposal costs. \emph{Recovery processing rate} (RPR) is
the nonscrap share, including refurbishment, partial recovery, and component
recovery. Demand coverage is outside this metric. \emph{Inspection cost savings} (ICS) is the cost of
full inspection on every arrival minus actual inspection cost.
SSADS--Keyword skip/quick/full splits are 67/33/0\% (S1), 58/0/42\% (S2), and
0/10/90\% (S3).
Across all 30 batches, the exact counts are 10,010/4,990/0,
8,644/0/6,356, and 0/2,931/27,069, respectively.

\subsection{Results}\label{sec:results}
Tables~\ref{tab:main} and~\ref{tab:calibration} report economic performance and the
Pearson correlation between each extractor's condition factor and latent yield,
respectively.

\begin{table*}[!t]
\caption{Net Recovery Performance Across Scenarios}
\label{tab:main}
\centering\footnotesize
\begin{tabularx}{\textwidth}{@{}l*{9}{Y}@{}}
\toprule
 & \multicolumn{3}{c}{S1: IT}
 & \multicolumn{3}{c}{S2: Aircraft}
 & \multicolumn{3}{c}{S3: Consumer} \\
\cmidrule(l{24pt}r{2pt}){2-4}%
\cmidrule(l{24pt}r{2pt}){5-7}%
\cmidrule(l{24pt}){8-10}
System & TRV & RPR & ICS & TRV & RPR & ICS & TRV & RPR & ICS \\
\midrule
Random routing & 333 & 67\% & 38 & 1{,}128 & 67\% & 50 & $-$5 & 66\% & 20 \\
FIFO routing & 614 &100\% & 38 & 1{,}233 & 61\% & 50 & 81 &100\% & 20 \\
\begin{tabular}[t]{@{}l@{}}Structured + noisy\\full inspection\end{tabular} & 405 & 26\% & 0 & 1{,}382 & 36\% & 0 & 70 & 90\% & 0 \\
\begin{tabular}[t]{@{}l@{}}Structured + semantic\\adaptive inspection\end{tabular} & 611 &100\% & 34 & 1{,}618 & 51\% & 29 & 71 & 95\% & 1 \\
Oracle full inspection & 406 & 26\% & 0 & 1{,}389 & 36\% & 0 & 70 & 90\% & 0 \\
No-signal / no-inspection & 614 &100\% & 38 & 1{,}676 & 61\% & 50 & 81 &100\% & 20 \\
\begin{tabular}[t]{@{}l@{}}Semantic-only /\\threshold routing\end{tabular} & 553 &100\% & 34 & 1{,}677 & 96\% & 29 & 42 & 64\% & 1 \\
\textbf{SSADS--Keyword}
            & \textbf{611} & \textbf{100\%} & \textbf{34}
            & \textbf{1{,}603} & \textbf{51\%} & \textbf{29}
            & \textbf{71} & \textbf{95\%} & \textbf{1} \\
\midrule
\begin{tabular}[t]{@{}l@{}}Lift vs. structured +\\noisy full inspection\end{tabular}
 & \multicolumn{3}{c}{$+206$~($+50.9\%$)}
 & \multicolumn{3}{c}{$+221$~($+16.0\%$)}
 & \multicolumn{3}{c}{$+1.28$~($+1.8\%$)} \\
\bottomrule
\end{tabularx}
\par\smallskip
\begin{minipage}{\textwidth}
\footnotesize\raggedright\emph{Note.}
Values are in thousands of United States dollars; means over 30 seeds, rounded to
the nearest thousand dollars except the S3 lift. Lifts over structured noisy full
inspection are significant (paired Wilcoxon $p{<}0.001$, $n{=}30$).
Paired-bootstrap 95\% intervals are S1 $[+49.2,+52.8]\%$, S2
$[+15.6,+16.4]\%$, and S3 $[+1.7,+1.9]\%$. These intervals and $p$-values
describe variation across simulated seeds, not field generalization. TRV seed standard deviations for
SSADS--Keyword are 44/80/2.5 and for structured noisy full inspection are
42/69/2.4. Bold marks SSADS--Keyword.
Reproduction command: \texttt{run\_summary.py --seeds 0-29}.
\end{minipage}
\end{table*}

\begin{table}[!t]
\caption{Extractor-to-Latent-Yield Correlation}
\label{tab:calibration}
\centering
\footnotesize
\setlength{\tabcolsep}{4pt}
\begin{tabular*}{\columnwidth}{@{\extracolsep{\fill}}lccc@{}}
\toprule
Scenario & Keyword & Phrase & LLM \\
         &          & matcher & DeepSeek \\
\midrule
S1: IT infra.  & 0.48 & 0.73 & 0.64 \\
S2: Aircraft   & 0.32 & 0.62 & 0.74 \\
S3: Consumer   & 0.36 & 0.74 & 0.80 \\
\bottomrule
\end{tabular*}
\par\smallskip
\begin{minipage}{\columnwidth}
\footnotesize\raggedright\emph{Note.} Keyword and phrase scores use the full
30-seed corpus. The DeepSeek diagnostic uses 150 cached records per scenario
(\texttt{deepseek-chat}, temperature 0). Values are correlations with latent
yield, not calibrated probabilities.
\end{minipage}
\end{table}

Table~\ref{tab:main} reports means over 30 paired seeds. Within a seed, every
method sees the same asset population, per-asset inspection perturbations, and
pre-drawn realized component outcomes.

\smallskip\noindent\textbf{Full-inspection comparison.}
SSADS--Keyword exceeds the structured noisy-full-inspection baseline by 50.9\% in S1,
16.0\% in S2, and 1.8\% in S3. The oracle row is close to structured, indicating
that observation noise contributes less than the labor consumed by inspecting every
arrival. SSADS saves \$34.3K, \$28.8K, and \$1.0K in inspection cost, respectively.
Its mean TRV per arrival is \$1,222/\$3,206/\$71 across S1/S2/S3.

\smallskip\noindent\textbf{Economic-only comparison.}
By avoiding all inspection cost, the no-signal / no-inspection policy has higher
simulated TRV than SSADS--Keyword in all three scenarios. Its advantage is
0.5/4.6/14.3\% for S1/S2/S3. The matched-cost analysis below holds inspection
expenditure fixed to isolate the value of semantic targeting.

\smallskip\noindent\textbf{Text, structured, and combined features.}
In the generator, latent condition is sampled independently of age, asset type,
and BOM counts; accordingly, the structured-only prior has near-zero correlation
with latent condition ($-0.01/0.00/-0.01$ for S1/S2/S3). Combining those features
with keyword outputs raises correlation from $0.48/0.32/0.36$ to
$0.71/0.60/0.53$. Structured + semantic adaptive inspection reaches
\$611K/\$1,618K/\$71.2K, improving over SSADS--Keyword by \$15.5K in S2 and
negligibly in S1/S3.

\subsection{Condition-Factor Extraction Accuracy}\label{sec:calibration}

Table~\ref{tab:calibration} reports Pearson $r$ between each extractor's $\phi$ and
the latent yield factor. Keyword and phrase results pool 15,000/15,000/30,000
records over 30
seeds; the cached DeepSeek correlation check uses 150 prespecified records per scenario.
Here, $\sigma$ measures note informativeness rather than probability, so expected
calibration error and Brier score are not applicable. Section~\ref{sec:error}
instead evaluates operational selectivity using high-score bad-skip rates and the
association between $\sigma$ and absolute condition error; $\sigma$-binned error
diagnostics are released with the benchmark~\cite{RLDBRepo}.

\subsection{Effect of Extractor Choice on Recovery Value}\label{sec:cal2val}
Section~\ref{sec:calibration} compares condition-factor accuracy across extractors. We
test whether that improvement translates into higher \emph{recovery value}. We rerun the full
seeded TRV pipeline with the phrase matcher (Pearson $r$ for S1/S2/S3:
$0.73/0.62/0.74$) in place of keyword ($0.48/0.32/0.36$), changing nothing
else. SSADS--Phrase reaches \$614K in S1, \$1,720K in S2, and \$80.9K in S3. The largest
change is S2, where it rises \$117.1K over SSADS--Keyword and exceeds the no-signal /
no-inspection comparator by 2.6\%.

SSADS--DeepSeek reaches \$604K in S1, \$1,703K in S2, and \$81.0K in S3.
Only S2 exceeds no-signal / no-inspection, by 1.6\%. The 150-record sample in
Table~\ref{tab:calibration} provides the correlation estimate, whereas these TRV
values use all 15{,}000/15{,}000/30{,}000 records. Stronger
extraction is most useful when text diversity and capacity constraints affect
ranking.

\subsection{Error Analysis and Case Studies}\label{sec:error}
The main failure mode is \emph{overestimation}: the policy skips
inspection on a high score ($\sigma{\geq}\tau_h$, $\phi{>}0.7$), but the true
yield is low ($<0.30$). Such scoring errors allow low-yield assets to bypass
inspection. Under the keyword reader, $4.8\%$ (S1) and $10.1\%$ (S2) of
high-score skips have true yields below $0.30$; no such cases occur in S3. The
$10.1\%$ S2 rate precludes autonomous use of the keyword reader in aircraft MRO.
Replacing keyword scores with the cached DeepSeek outputs in the full decision
pipeline does not increase this failure rate. DeepSeek assigns high $\sigma$ to a
smaller, more selective set of notes, reducing the rate to $2.7\%$ (S1) and
$0\%$ (S2; none of the $2{,}675$ skipped assets has a true yield below $0.30$;
\texttt{run\_diagnostics.py}).
Across keyword assets, $\sigma$ correlates negatively with absolute condition error
($-0.52/-0.65/-0.36$), so higher scores are generally more selective. The released
diagnostics link every low-yield skip to disposition and realized value. In S2, 333 of
421 low-yield skips under the keyword reader enter component recovery and 88 are
capacity-scrapped; 35
(8.3\%) have negative simulated net value. Their aggregate net value is still
positive because high component prices reward low-yield recovery.

\subsection{Sensitivity Analysis}\label{sec:sensitivity}
At half labor capacity, SSADS--Keyword TRV falls 0.6\% in S1 and 49.9\% in S2;
the prespecified S3 inspection plan plus default scrap handling exceeds the
reduced 80-hour budget and is infeasible. The prespecified defaults serve as fixed
reference settings. Nearby threshold settings improve TRV
by at most 0.1/4.3/3.4\%
for S1/S2/S3.
With latent assets and noise draws paired across variants, changing omission
(0--30\%) and severity perturbation (0--40\%) changes TRV by at most
0.19/2.75/0.42\% in S1/S2/S3.

\smallskip\noindent\textbf{Held-out vocabulary family.} To test sensitivity to an
unseen vocabulary family, we remove one family at a time from the keyword reader
with thresholds fixed. Removing the negative
vocabulary \emph{inverts} the condition signal, with $r$ falling from $0.48$ to
$-0.24$ in S1 and $0.32$ to $-0.09$ in S2, because unmatched damaged notes fall
back to the maximally optimistic $\phi{=}1.0$ prior. TRV still moves by only
$-0.3\%$ (S1) and $-6.9\%$ (S2), and the S2 bad-skip rate is unchanged at
$10.1\%$. The weak TRV response to an inverted signal motivates a conservative
fallback prior for unmatched notes.

\subsection{Matched-Cost Targeting Ablation}\label{sec:ablation}
For each seed, matched-cost random assigns exactly the same skip/quick/full counts
as SSADS, but shuffles them across assets. Score-guided targeting changes mean TRV by
\$0 in S1, $+\$53.9$K in S2 ($p{<}0.001$), and $+\$18$ in S3
($p{<}0.001$ but economically negligible). In these scenarios, targeting creates
material economic value only in S2, where the allocator processes about half of the
arrivals; capacity also binds in S3 without a material gain. Removing
margin-per-minute ranking from SSADS, while holding its inspection policy and
action set fixed, reduces TRV by
0.0/38.0/5.6\% for S1/S2/S3, reinforcing
that allocation is most consequential in aviation.

\section{Discussion}

\subsection{Practical Implications}
The framework requires an analytical allocator and an extractor with a two-output
interface. A practical deployment can begin with inspection prioritization, with
required checks retained and suggested depth, observed condition, and outcomes
logged for local threshold estimation. In S2, the workflow prioritizes uncertain
records for inspection, reserves processing labor for high-margin parts, and retains
qualified inspection and authorized return-to-service sign-off under applicable
repair-station procedures~\cite{eCFRPart145}.

\smallskip\noindent\textbf{Extractor selection.}
The deterministic phrase matcher reaches Pearson $r{=}0.62$--$0.74$ with exact
reproducibility. In S2, SSADS--DeepSeek reaches \$1,703K, below SSADS--Phrase at
\$1,720K. The comparison illustrates that
extractor choice depends on downstream recovery performance as well as privacy,
auditability, latency, and model-version stability. The common interface leaves the
downstream decision model unchanged.

\subsection{Limitations}\label{sec:limits}
\noindent\textbf{Synthetic scope.}
RLDB uses stylized prices, yields, capacities, inspection noise, and fixed return-note
templates to isolate the decision mechanism. This controlled design supports
reproducible analysis of how semantic signals change inspection and allocation,
rather than an estimate of field effectiveness. The gains from phrase and DeepSeek
readers identify extractor robustness as an important direction. Further evaluation
can include broader note families and local calibration of $\sigma$.

\smallskip\noindent\textbf{Economic objective.}
TRV omits the costs of certification errors, latent safety failures, warranty
exposure, and most downstream failures. These omissions can favor risk-blind
no-inspection policies and preclude interpreting unsafe S2 skips as financially
optimal. In S3, configured rework is realized only after allocation. It is neither
forecast nor reserved in the first-stage capacity plan.

\subsection{Benchmark and Reproducibility}
RLDB provides configurations, noisy-note generators, seven comparators, exact prompts,
DeepSeek score caches, paired seeds, and table scripts in the companion
repository~\cite{RLDBRepo}. The repository includes data profiles, parameters,
per-asset values, sensitivity, matched-cost, selective-risk, and exact-allocation
results. The released reproduction uses Python 3.12.13 with
pinned direct dependencies and requires no graphics processing unit (GPU); a
released DeepSeek cache covers every unique note used in the evaluation and replays
parsed scores without a live call. The cache fixes the LLM scores used in the
reported evaluation. A manifest records the commands and the Secure Hash Algorithm
256-bit (SHA-256) digests of the audit tables and score caches. The reproduction
verifier byte-compares six console outputs and 16 audit tables against their
released references. The original API date, provider snapshot, token usage, cost,
latency, retries, and live parse-failure count were not recorded and cannot be
reconstructed from the cache; these deployment attributes are not evaluated.

\section{Conclusion}

SSADS uses return notes to update expected condition and allocate inspection effort
before recovery decisions are made. In the RLDB evaluation, the keyword
implementation improves net recovery value relative to noisy full inspection while
reducing inspection cost. The matched-cost analysis shows that semantic targeting
is materially valuable in the aircraft scenario but negligible under the
information technology and consumer configurations. Phrase and DeepSeek
extractors further improve the aircraft result, indicating that the modular design
can benefit from stronger semantic readers. A prospective advisory deployment
should evaluate the approach on operational notes while retaining mandatory safety
inspections and human approval.

\IEEEtriggeratref{13}

\end{document}